\documentclass[final,authoryear,5p,times, twocolumn]{elsarticle}

\usepackage{multirow}
\usepackage{booktabs}
\usepackage[table,xcdraw,dvipsnames]{xcolor}

\usepackage{lineno}
\usepackage{graphicx}
\usepackage{color}
\usepackage{amssymb}
\usepackage{amsmath}
\usepackage{bigints}
\usepackage[bookmarksdepth=3]{hyperref}
\usepackage{lineno}
\usepackage{mathtools}
\usepackage{blkarray}
\usepackage[capposition=bottom]{floatrow}
\usepackage{varwidth}
\usepackage{cuted}

\usepackage{pifont}
\graphicspath{{img/}}
\usepackage{booktabs}

\usepackage{pgf,tikz}
\usepackage{mathrsfs}
\usetikzlibrary{arrows}
\hypersetup{
    colorlinks=true,
    citecolor=red,
    linkcolor=blue,
    filecolor=blue,      
    urlcolor=red,}
    
\journal{CS294-026: Computational Photography and Computer Vision}

\begin{document}

\begin{frontmatter}

\title{Jenga Stacking Based on 6D Pose Estimation for Architectural Form Finding Process}

\author[add1]{Zixun Huang}

\address[add1]{University of California, Berkeley, USA}

\begin{abstract}
This paper includes a review of current state of the art 6d pose estimation methods, as well as a discussion of which pose estimation method should be used in two types of architectural design scenarios. Taking the latest pose estimation research Gen6d as an example, we make a qualitative assessment of the current openset methods in terms of application level, prediction speed, resistance to occlusion, accuracy, resistance to environmental interference, etc. In addition, we try to combine 6D pose estimation and building wind environment assessment to create tangible architectural design approach, we discuss the limitations of the method and point out the direction in which 6d pose estimation is eager to progress in this scenario.

\end{abstract}

\begin{keyword}

Computational Design; Tangible Feedback; Discrete Structure; Augmented Reality; CFD Simulation

\end{keyword}

\end{frontmatter}

\section{Introduction}
\subsection{Background}

\begin{figure}[h]
    \includegraphics[width=1\textwidth]{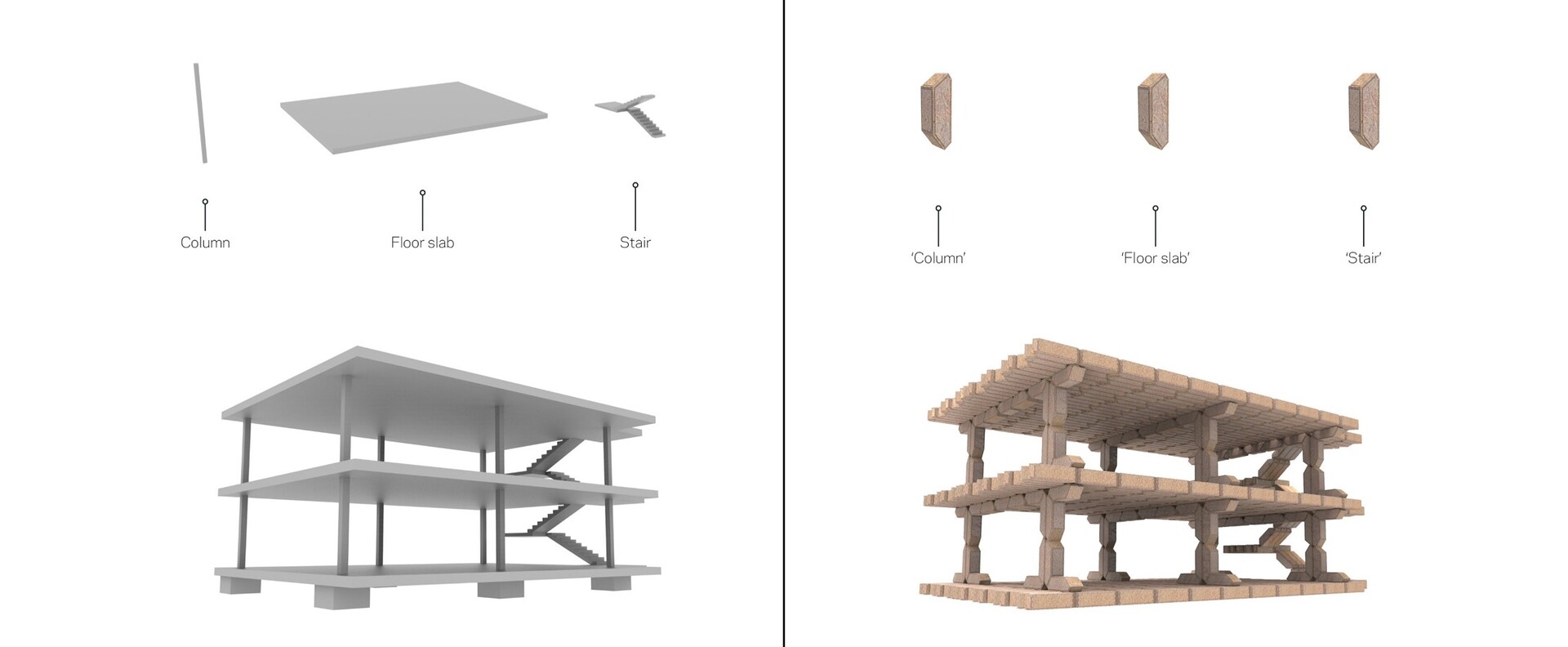}
    \caption{Discrete parts as a reassembly of Maison Dom-ino. Image: Ivo Tedbury, Semblr, Unit 19/DCL, 2017.}
    \label{F1}
\end{figure}

The splines, pixels, voxels, and bits associated with digital innovation and technological progress in the last several decades have enabled architectural form to be re-conceptualized not as static, permanent objects but as part of a larger network of data and communication between different kinds of architectural systems in a continuous state of flux \citep{carpo2014breaking}. The mass standardization of Maison Dom-Ino (Fig\ref{F1}) transformed the production of the post-war environment. Today, technologies of mass customization from the “first digital turn” make it so that architects no longer have to think of space as being constructed of fixed elements or objects \citep{carpo2012digital}. With a discrete approach, the dichotomy between the virtual—the way things are designed—and the physical—the way things are realized—becomes much smaller. Methods and processes of design, fabrication, and assembly become more streamlined. The role of the architect becomes less concerned with the final form a building may take, and instead engages more with the overarching economic, social, material, and technical framework in which it is produced.\par

\begin{figure}[h]
    \includegraphics[width=1\textwidth]{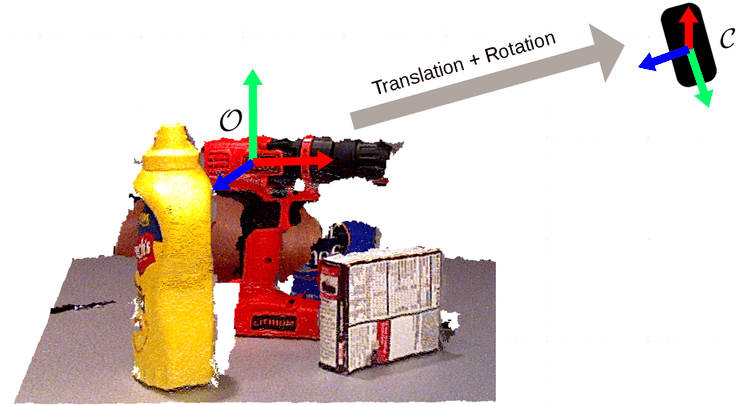}
    \caption{The goal of 6D pose estimation is to find the translation and rotation from the object \citep{gao2018occlusion}}
    \label{F2}
\end{figure}

Object Pose Estimation is a typical task in computer vision that consists of identifying specific object instances in an image and determining each object’s position and orientation relative to some coordinate system (Fig\ref{F2}). This technology is well suited to the needs of the discrete architectural design process. Architectural design is inevitably dependent on the manipulation and adjustment of manual models. The design of a discrete building is like a game of building blocks, where geometric elements with the same or different architectural functions are put together and combined.\par

\subsection{Objective}
This project is a response to the theory of discrete architecture in the context of computer vision, trying to provide a possible technological development for this bold and avant-garde architectural theory. The project is based on Jenga's development of a game process that can interact with realistic scenes, i.e., locating the placement of wooden blocks through 6DoF, and then providing feedback on the design process through existing information, such as generating and suggesting the placement of the next block, generating and visualizing wind field predictions for the current design, etc.\par

\subsection{Problem Statement}
Most of the current digital construction research tends to focus too much on the geometric degrees of freedom offered by the construction method itself, following predetermined forms and paths in practice \citep{xu2020robotic}, often neglecting to respond to the design process itself. Generative design based on semantic \citep{schumacher2009parametricism, schumacher2011autopoiesis} or physical simulations \citep{roudsari2013ladybug}, on the other hand, relies heavily on the computing power and neglects the experience and inspiration that hand models bring to the designer. There are also studies that evaluate architectural designs by producing physical devices \citep{song2021research}, but such methods are costly and have insufficient generalization capabilities, which can pull down their ability to update and iterate the architectural design process.\par

\subsection{6D Pose Estimation}
In the literature, algorithms for 6D object pose estimation can be divided into three main categories: Direct methods \citep{dttd2, hu2020single, xiang2017posecnn, tekin2018real} are used to regress the pose of 6D objects directly from images or depth map. For example, PoseCNN \citep{xiang2017posecnn} directly and separately regress rotation and translation after segmentation.\par
In Perspective-n-Point based method \citep{peng2019pvnet, rad2017bb8}, 2D features and 3d correspondences are extracted,  Perspective-n-Point (PnP) problem can be solved to obtain the pose of the object with respect to the camera coordinate frame. These methods mainly focus on the feature extraction but usually cannot work without models. Representatively, PVNet \citep{peng2019pvnet} regresses pixel-wise vectors pointing to the keypoints and use these vectors to vote for keypoint locations. \par
Template-matching based approaches \citep{sundermeyer2018implicit, wohlhart2015learning, sundermeyer2020multi} provide a series of support images with pose information and find the closest images to the query image and match them. They usually embed images into feature vectors and then compute similarities using distances of feature vectors.\par
Another important concept in pose estimation is open set or close set. Open set methods can be generalized to unseen objects and classes, while close set cannot. In general, direct regression is limited by the size of the dataset and currently cannot train models that can be generalized to openset. However, there are synthetic datasets \citep{he2022fs6d} that can help to approach this. Image matching, on the other hand, has a good performance in generalization.

\subsection{Pose Estimation in Architectural Scenarios}
Discrete and prefab is currently an important architectural design trend in the discipline of architecture \citep{smith2010prefab}. Modularity helps to increase the flexibility and scalability of architectural design, reduce costs, and improve sustainability. In this discrete context, there are two situations where the intervention of pose estimation is urgently needed.\par
\begin{figure}[h]
    \includegraphics[width=1\textwidth]{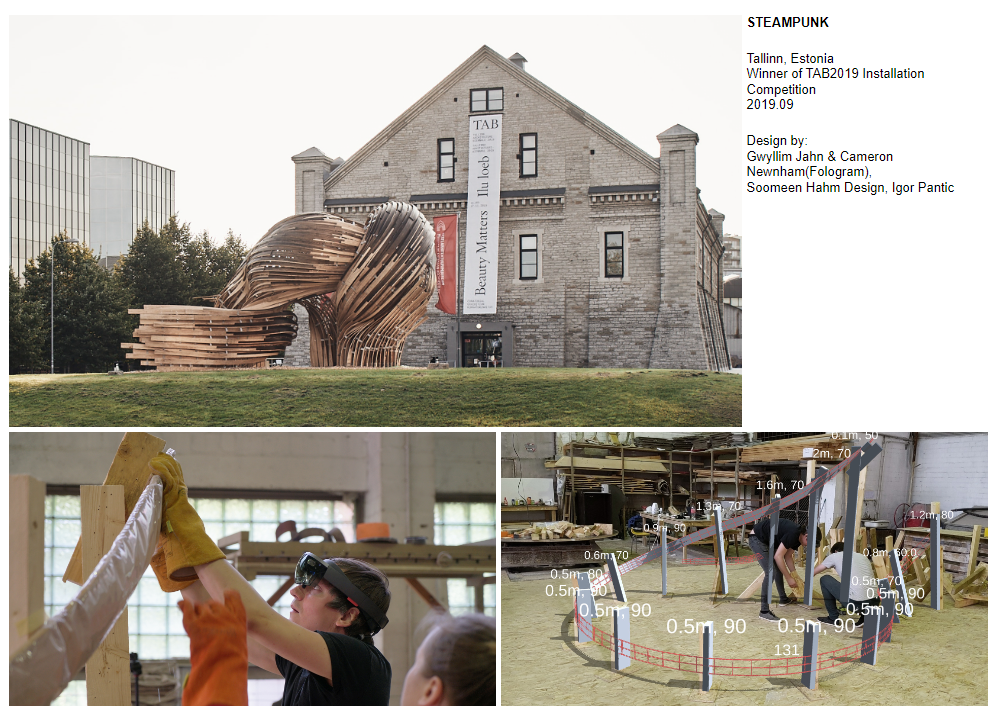}
    \caption{G. Janh, C. Newnham, S. Hahn, I. Pantic, Steampunk Pavilion, Fologram, (last modified October 2019). https://vimeo.com/365917769, 2019.}
    \label{F3}
\end{figure}

One of the main uses of pose estimation is to help locate objects in AR scenes, while digital architectural designs \citep{kyjanek2019implementation, sun2018hybrid, wibranek2019digital, song2020bloomshell} in discrete contexts are also beginning to use AR to assist in construction. The Steampunk project (Fig\ref{F3}) is a strong demonstration of the power in applying AR technology to field construction projects to aid in the positioning of discrete structures. But in fact the power of AR and computer vision is far from being fully explored in that process. In many real-world construction processes, the errors in the installation of modules are not negligible due to environmental factors and material properties, etc. At this time, designing flexible and variable unit structures and using pose estimation to track the actual position of modules in the real world can help generate and adjust the position and orientation of the next module in real time, thus greatly improving the flexibility of modular construction. 

A very small amount of cutting-edge research \citep{sun2018hybrid, song2020bloomshell} has gradually begun to focus on the feedback of the physical construction process on architectural design in AR. For example, the study \citep{sun2018hybrid} uses the iterative closest point (ICP) to evaluate the fitness level between the installed surrounding bricks and the alternative brick from the pool. However, these studies usually focus on the whole iterative process of design and construction, while the algorithms used are usually only brought in one sentence. This paper advocates the fitness of algorithms and application scenarios, focusing more on how different types of algorithms perform in this process, such as what advantages exist in the algorithms used and what disadvantages exist in which scenarios, in an attempt to introduce the most suitable algorithms to designers and to inspire developers on the directions that need to be improved in this research area.\par
In addition, the new materialism promotes the development of architectural design in the direction of performance optimization, including structural topology optimization \citep{zhou2018ameba}, wind environment and thermal environment simulation \citep{roudsari2013ladybug}. However, this series of researches has also brought the process of architectural design into the virtual space. It is undeniable that for most architects, manipulating manual models is still their main source of inspiration, and we cannot ignore the fact that the touch of the hand on materials is an important part of architectural design.\par
Computer vision and reality augmentation technologies bring the opportunity to combine physical simulation and handmade models. Through pose estimation, we can allow computers to perceive how building modules are posed and composed in space, so that physical simulations and even real-time calculations can be performed in digital space, and the visualization of the physical simulation can be superimposed on the handmade model through AR technology.\par

In the design phase, we should use pose estimation methods that have openset, model-free, and other properties that do not limit the flexibility of the design as much as possible. The building components designed by architects are usually not the objects we see everywhere in our daily lives, and there is a wide range of scales in the handmade models used for architectural design. This points our focus to the openset.\par

In addition, although we usually produce fine CAD models during the architectural design process, most of these models will not be attached with very realistic materials during the refinement process. This means that the only model information available for pose estimation is often geometric information, and that geometric information is constantly adjusted as the design process progresses. Here, we advocate the use of a pose estimation method that does not rely on the model and depth map, but only on RGB information.\par

Instead, during the construction process we should use methods that rely on high precision CAD models and close sets. At that stage, most of the design has been finalized and the time cost of training the computational model for the close set is very small compared to the cost of field construction. The input of more 3D information and targeted adjustment of model parameters can help to improve the accuracy and speed of pose estimation during construction.\par


\section{Methodology}

\subsection{Gen6D}
\begin{figure}[h]
    \includegraphics[width=1\textwidth]{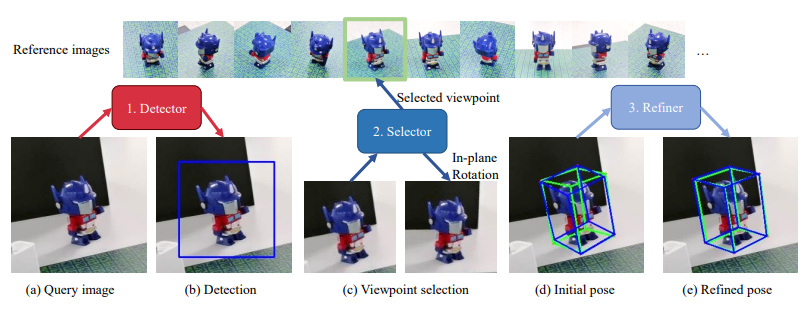}
    \caption{Gen6D consists of a detector which detects the object in the query image, a viewpoint selector which selects the most similar viewpoint from reference images, and a pose refiner which refines the initial pose into an accurate object pose. \citep{liu2022gen6d}}
    \label{F4}
\end{figure}

As an example, this year's latest study Gen6d \citep{liu2022gen6d}, which is based on openset and model-free setup, achieves competitive results compared to other state-of-the-art pose estimators. In this paper, the method is tested at the application level of assisted architectural design, and the remaining bottlenecks and development directions are presented. Gen6d utilizes an image matching approach that can be extended to openset and is not dependent on CAD models. This pose estimator (Fig\ref{F4}) consists of a detector which detects the object in the query image, a viewpoint selector which selects the most similar viewpoint from reference images, and a pose refiner which refines the initial pose into an accurate object pose.\par

Here are some tips this pose estimator used. Since it is not rely on model, even not rely on depth map. They need to find a way to calculate the depth of object center to estimate translation. They normalize all objects into a unit sphere. So that Detected bounding box size can be used in computing the depth of the object center. Additionally, they applied in plane rotation to every reference images, so that there can be more opportunities to match query image with support images. And they also applied a transformer and attention layers to the network, so that the selector can share information across reference images.

\subsection{Experiment Design}
\begin{figure}[h]
    \includegraphics[width=1\textwidth]{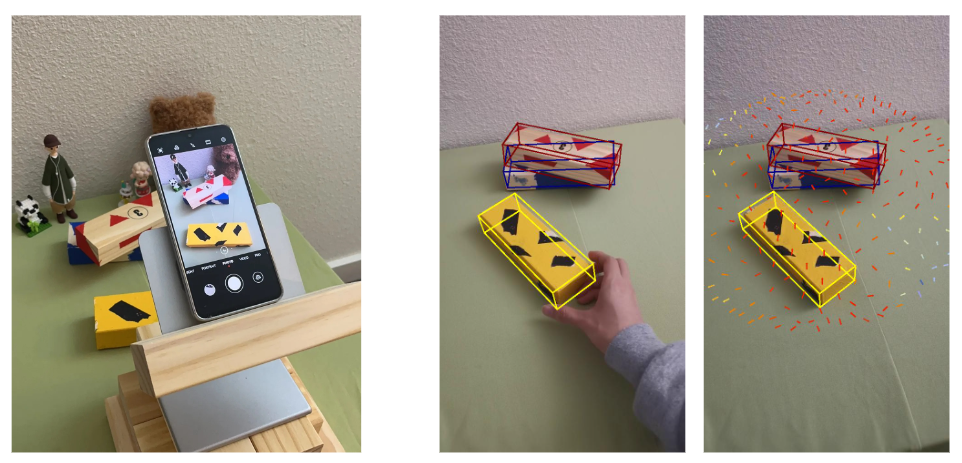}
    \caption{My experiment setup \citep{liu2022gen6d}}
    \label{F5}
\end{figure}
In this project, we try to combine tangible physical models and CFD predictions by using 6D pose estimation as a bridge. For my experiment (Fig\ref{F5}), we used three blocks of the same size and randomly attached textures (with 3 different colors: blue, yellow, red; since gen6d works better when texture-less objects are removed). The purpose of this experiment is to track the pose of wooden blocks while the camera is fixed.\par

\subsection{Pre-processing}
\begin{figure}[h]
    \includegraphics[width=1\textwidth]{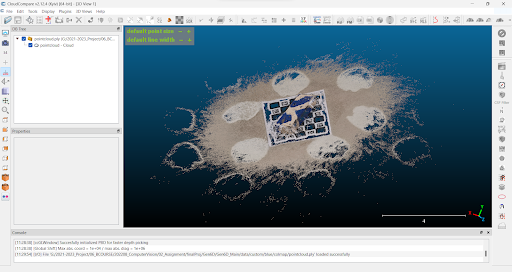}
    \caption{Point cloud reconstruction of the blue block before being cropped.}
    \label{F6}
\end{figure}

\begin{figure}[h]
    \includegraphics[width=1\textwidth]{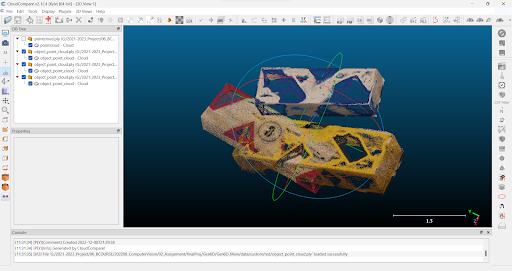}
    \caption{Point cloud reconstruction of 3 blocks after being cropped. They may have different sizes before normalization.}
    \label{F7}
\end{figure}

First we need to do a rough reconstruction for object normalization. Take the blue wood block as an example, we record a reference video around it from different angles. Ffmpeg is used to break videos down into images and input them to COLMAP \citep{schonberger2016structure, schoenberger2016mvs} for point cloud reconstruction by the structure from Motion method. Then the point cloud needs to be cropped manually to remove the environment (Fig \ref{F6}, \ref{F7}). With the point cloud, we can calculate the center and size of the object, which helps us to normalize the object to a unit sphere. In addition, in this process, we ensure that 1. the object in the reference video is static; 2. there is enough texture in the reference video for COLMAP to extract features for matching. \par
In addition, we need to manually record the x-direction and z-direction of the bounding of point cloud, this reconstruction process can help us to label each reference image with extrinsic matrix i.e., the transformation of the object from world coordinates to the camera coordinate system. \par
After that, I placed the blocks in sequence on the table and kept adjusting the position.
The pose estimator calculates the pose of each block at each frame and provides this data to other design aids, such as CFD.

\section{Result}
\begin{figure}[h]
    \includegraphics[width=1\textwidth]{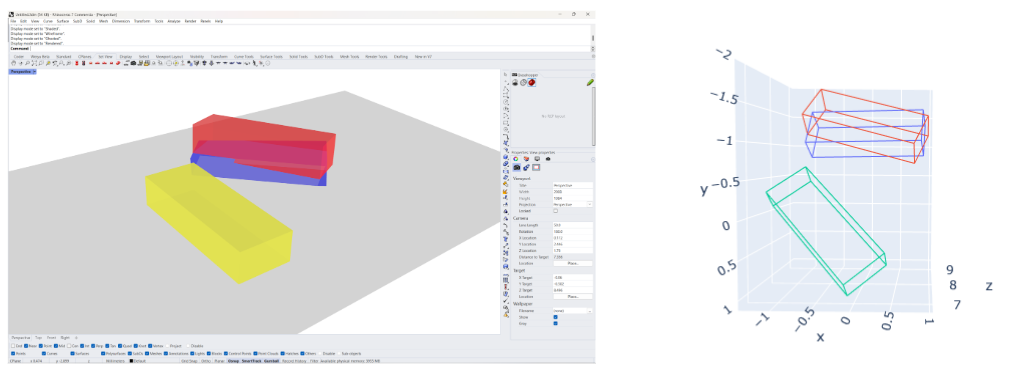}
    \caption{the wooden blocks with predicted poses in our modeling space}
    \label{F8}
\end{figure}

In the test, we successively put blue, red and yellow blocks on the table. And the 6D pose of the blue, red and yellow wooden blocks are predicted respectively. The model can accurately track the spatial position of the blocks. With the predicted extrinsic information, we can reproduce the relative positions of the wooden blocks into our modeling space (Fig \ref{F8}). For example picking up into modeling software, or visualizing that 3D information through python open3d.\par
\begin{figure}[h]
    \includegraphics[width=1\textwidth]{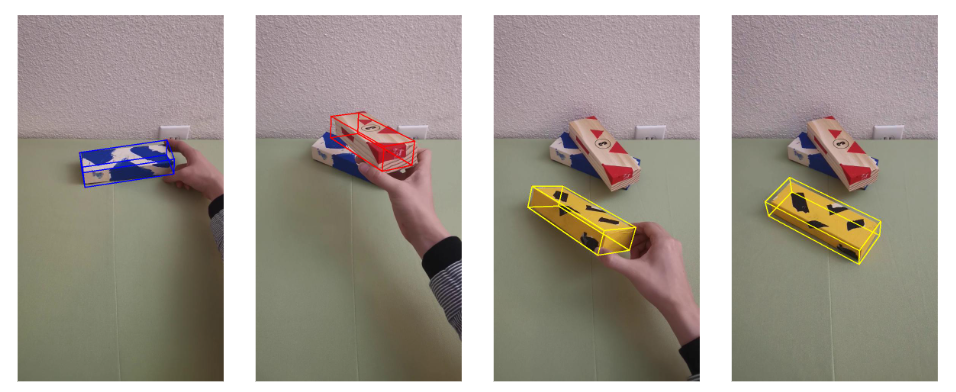}
    \caption{Result of Experiment 1}
    \label{F9}
\end{figure}

Interestingly, during my experiments, we found that prompt tuning can improve the accuracy of the model prediction, due to the fact that the model borrows the transformer and attention layer in the network structure which are widely used in Neural Language Processing. Specifically, in prediction, we need to input the position of the previous frame. So we can adjust the prediction result by the method of prompt tuning to make it more accurate in the place where the prediction error looks large through following steps: 1. predict the frames in a small range, the accuracy will be gradually increasing as the block moves. 2. input the position of the last frame. Re-predict frames from the first again, so that the overall prediction result can be improved. Here are the final predicted results of this experiment: Fig \ref{F9}.
\subsection{Existing Problem}
\begin{figure}[h]
    \includegraphics[width=1\textwidth]{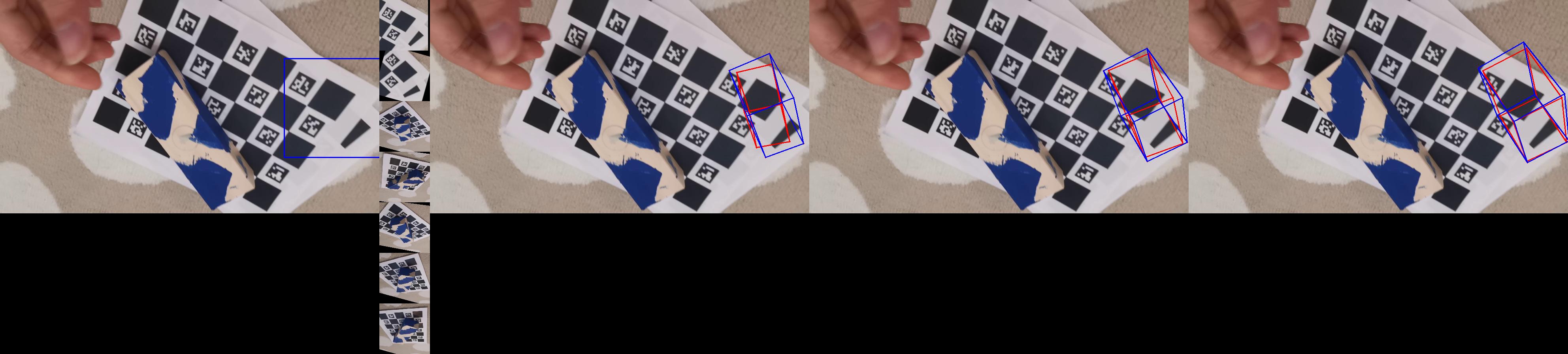}
    \caption{Detector affected by surrounding textures.}
    \label{F10}
\end{figure}
Here are some existing problems:
\begin{enumerate}
    \item The prediction process is easily affected by the texture of the surrounding environment objects. (Fig \ref{F10})
    \item The image matching method cannot effectively deal with occlusion, but since our camera is stationary in this experiment, there is only one wooden block being moved at the same time, so we can replace the object's pose after it is occluded with its last pose before it is occluded.
    \item When using prompt to update the pose prediction value, the process cannot be real time.
\end{enumerate}
\subsection{Extra Experiment: When Camera Moves}
But how are we going to handle the pose of the obscured object when the camera is moving. This scenario is also very common when we deal with architectural models, i.e. modules are usually easily obscured by another module, and we need to reproduce the positions of all modules in order to reconstruct the whole building in the modeling space.\par
For this experiment, we use the same setup as in the previous one, but this time our camera is movable. To simplify the problem, we assume that the objects do not move when the camera moves. In this case, we can still compute the pose of the occluded object B by predicting the pose of the unobstructed object A.\par

For example, blue is blocked, red is visible and pose is predictable, we can get the position of blue before it is blocked as its initial RT, in the process of camera movement afterwards, as the relative positions of red and blue blocks do not move, the difference of their pose compared to the initial state is the same. So we can compute the pose of the blue block in any following frames: $R_i^{blue} = R_i^{red} · (R_0^{red})^T · R_0^{blue}$.\par

\begin{figure}[h]
    \includegraphics[width=1\textwidth]{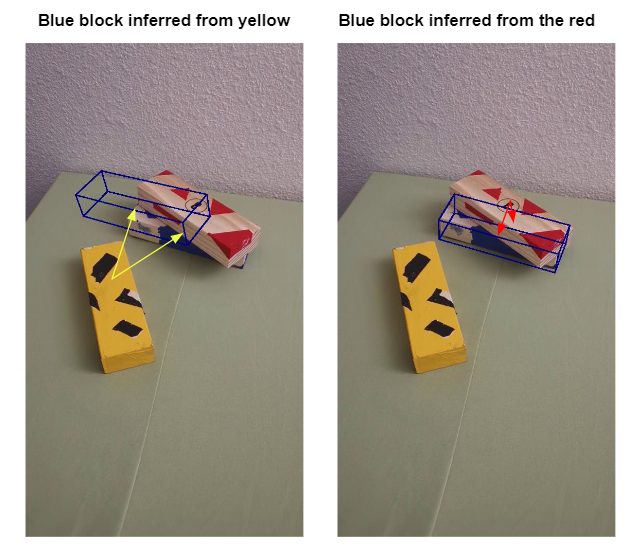}
    \caption{the rotation error in the prediction of the blue block when being inferred from the red or yellow.}
    \label{F11}
\end{figure}

This calculation method can help us to infer the pose of the occluded object in a specific case. However, its problem is that the rotation error in the prediction process will be magnified by the infer process. For example, we can infer the position of the blue block based on the red block or the yellow block, but the position inferred based on the red block will be more accurate because when there is a small error in the rotation prediction of the red block or the yellow block, the inferred value of the blue block will have a significant rotation around that error value, and the larger the distance, the more obvious that error will be. (Fig \ref{F11}).\par

\begin{figure}[h]
    \includegraphics[width=1\textwidth]{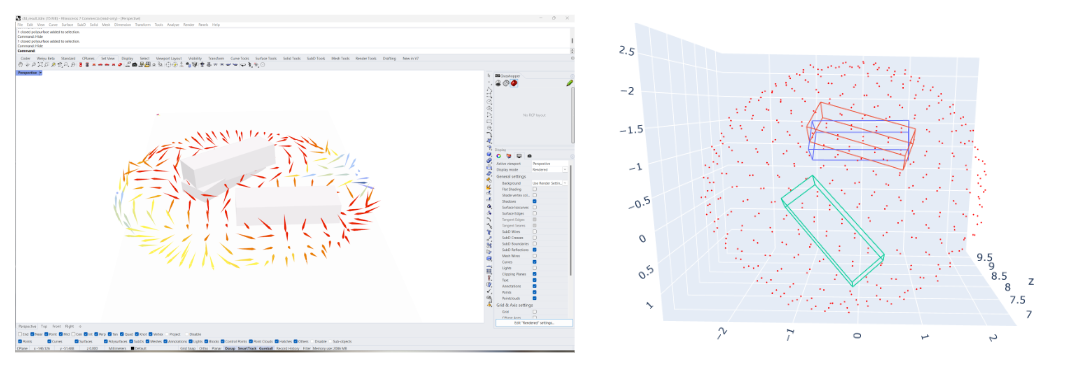}
    \caption{The wind environment prediction in modeling space.}
    \label{F12}
\end{figure}

\begin{figure}[h]
    \includegraphics[width=1\textwidth]{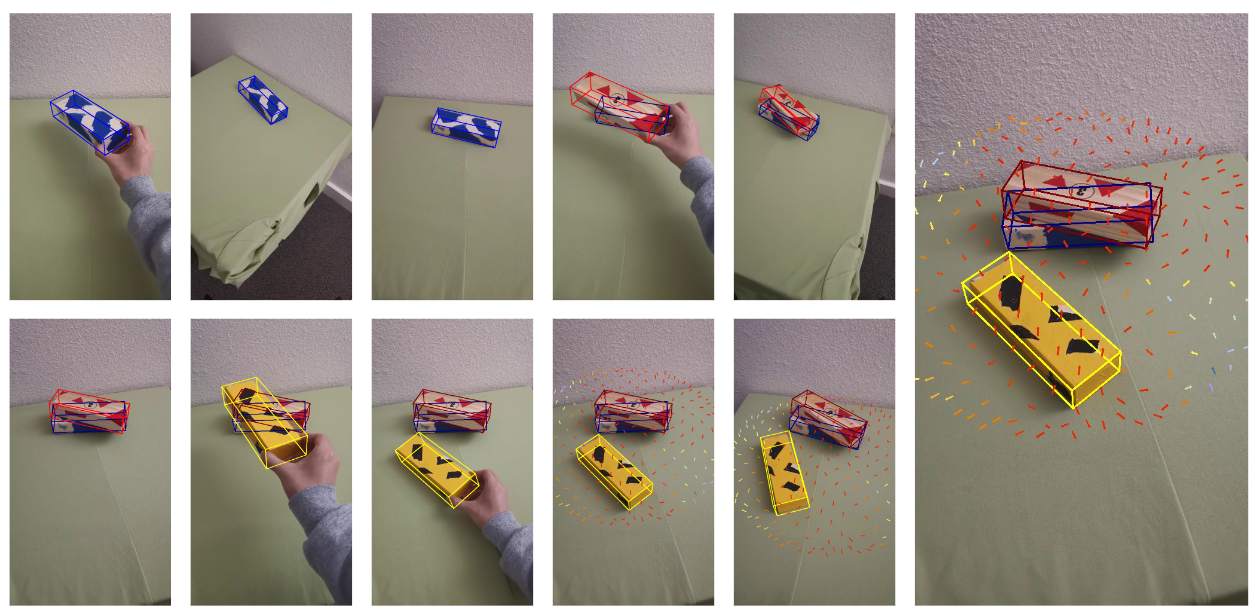}
    \caption{the results of the second experiment and CFD overlaying}
    \label{F13}
\end{figure}

Then we tried to combine 6D pose estimation and wind environment prediction to create tangible ways of building environment simulation. So that users can move wooden blocks and visualize the wind field by AR. The wind environment was predicted using BlueCFD after the relative poses are recovered into modelling space (Fig \ref{F12}). using the technique the same as in the second experiment, we overlay the CFD prediction into the frames with camera moving. Here are the results of the second experiment and CFD overlaying: Fig \ref{F13}.
\par


\section{Conclusion and Future Trends}
This paper demonstrates the feasibility of projecting design feedback into a manual model using pose estimator. The computer vision approach is applied to architectural design and responds to discrete architectural theories. We also dig into some problems that still need further improvement, such as occlusion, and filtering for cluttered environments.

Although there have been many studies \citep{xiang2017posecnn, peng2019pvnet} using methods such as Hough voting to make pose estimator to handle occlusion, occlusion may still be a problem to be solved in the case of openset. In addition, we will try to predict the multi-object pose directly \citep{irshad2022centersnap} in the application, which will be more accurate than deducing the pose from other objects. Using transformer can help the estimator to track the movement of the object, but we may need to remove the effect of prompt tuning on accuracy, as it makes it impossible to achieve real-time applications.

\section*{}
\bibliographystyle{apalike}
\bibliography{refs}

\end{document}